# An Algorithm for Automatically Updating a Forsyth-Edwards Notation String Without an Array Board Representation

Azlan Iqbal
*College of Computing and Informatics*
*Universiti Tenaga Nasional*
Putrajaya, Malaysia
azlan@uniten.edu.my

***Abstract*—We present an algorithm that correctly updates the Forsyth-Edwards Notation (FEN) chessboard character string after any move is made without the need for an intermediary array representation of the board. In particular, this relates to software that have to do with chess, certain chess variants and possibly even similar board games with comparable position representation. Even when performance may be equal or inferior to using arrays, the algorithm still provides an accurate and viable alternative to accomplishing the same thing, or when there may be a need for additional or side processing in conjunction with arrays. Furthermore, the end result (i.e. an updated FEN string) is immediately ready for export to any other internal module or external program, unlike with an intermediary array which needs to be first converted into a FEN string for export purposes. The algorithm is especially useful when there are no existing array-based modules to represent a visual board as it can do without them entirely. We provide examples that demonstrate the correctness of the algorithm given a variety of positions involving castling, en passant and pawn promotion.**

## I. Introduction

The Forsyth-Edwards Notation (FEN) is a method of describing a chess position using a string of characters. It was published in 1883 by Scottish newspaper journalist, David Forsyth (1854-1909), who emigrated to New Zealand and served as chess editor of the Glasgow Weekly Herald [1-2]. David Forsyth was also one of the compilers of an early regular column for the game of Go in a newspaper (Otago Witness, from February 1902 until March 1903), according to the New Zealand Go Society [3].

Originally just ‘Forsyth notation’, it was later known as ‘Forsyth–Edwards Notation’ when extended to support use with computers by the American computer scientist and programmer, Steven J. Edwards (1957-2016) [3-4]. We could not determine precisely when this occurred but presumably it became established after 1996 since both the 1992 and 1996 editions of [5] lists it as ‘Forsyth notation’ [6]. Appendix A provides more historical information on this matter to the interested reader.

## II. FEN Explained

The present format of the FEN is not very different, in principle, compared to its original form (see Appendix A). The board, with its vertical ‘files’ labeled ‘a’ through ‘h’ (from left to right) and horizontal ‘ranks’ numbered ‘1’ through ‘8’ (bottom to top) form a kind of Cartesian coordinate system that is read from the upper left to the lower right, one square at a time (i.e. *a8, b8, c8… a7, b7, c7… f1, g1 and h1*). Every time a white piece is encountered, a capital letter is used, i.e. ‘Q’ for queen, ‘R’ for rook, ‘B’ for bishop’, ‘N’ for knight and ‘P’ for pawn. If a black piece is encountered, the lower case alphabet is used. Empty squares are simply numbered; so a white queen and rook separated by one square would be presented as ‘Q1R’ and if by two squares, ‘Q2R’. When one rank is complete a slash (‘/’) is used to separate it from the next rank. After the last square in the last rank (i.e. rank 1) is described, there is a space followed by a ‘w’ or ‘b’ to indicate whose side it is to move.

This is followed by another space and the castling permissions. Given the rules of chess (some cases involving retrograde analysis [7-8] aside), it is difficult, if not impossible, to determine from a position alone if castling is still permissible because the relevant king or rook could have moved and returned to their original square, making castling now illegal. So if White may still castle on the king’s side or queen’s side, an upper case ‘K’ or ‘Q’ is used, respectively. For black, lower case ‘k’ or ‘q’. This string of up to four consecutive characters is applied only if castling of those types is allowed in the given position. If no castling permissions are allowed, as is the case for most positions, a single dash is used in its place. A blank space follows this as well.

After that is the possibility of en passant (or ‘in passing’) capture. This rule was reputedly added to chess, at least in some parts of the world, around the 15th century. It occurs when a pawn may capture another pawn that moved two squares forward in the previous move as if it had only moved one square. The relevant notation would be the target square of the capture, i.e. ‘behind’ the enemy pawn and described such as ‘*c6*’. Otherwise, a single dash is used. A pawn simply having moved two squares still warrants this part of the notation regardless if there is an enemy pawn that can make the capture on the next move.

This is followed by another space and the ‘halfmove clock’ which is the number of plies (or half-moves) since the last pawn advance or capture. It is useful when trying to determine a draw under the 50-move rule [9]. When unknown, it may be set to ‘0’. Finally, another space and the ‘fullmove number’ is indicated. This is incremented only after Black’s move or set to ‘1’ by default for composed positions with White to play; more details can be found in [10]. Fig. 1 shows an example (composed) position with its corresponding FEN string of characters as just described. White is to play (indicated by the lower case ‘w’).

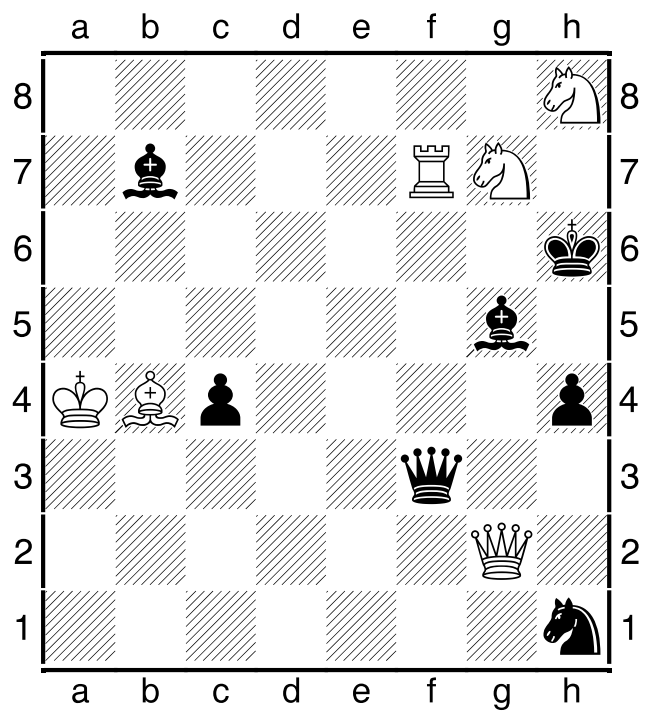

7N/1b3RN1/7k/6b1/KBp4p/5q2/6Q1/7n w - - 0 1

Fig. 1. Example chess position with its FEN.

The board itself is represented using a continuous sequence of characters including the slashes to separate the eight ranks. Then we have the first space and a 'w' indicating that White has the move. Castling is simply not possible (hence the first dash) and neither is an en passant capture (hence the second dash). Since it is a composed position with no known previous moves, the 'halfmove clock' is marked as '0'. The 'fullmove number' is marked as '1' given that White is to play. A FEN string is therefore very useful when representing a chess position (or 'node') in computers. The part of the string starting with 'w - - 0 1' is reputedly attributable to Steven J. Edwards as well [6].

Even the contemporary FEN format is lacking in terms of being able to determine, for example, whether there was a draw by threefold repetition or whether a draw offer was accepted. For such cases, a different format known as 'Extended Position Description' (EPD), developed in 1993 by John Stanback and the aforementioned Steven Edwards, may be used [11]. EPD is beyond the scope of this article, however. The FEN is nonetheless sufficient for describing standalone positions (not game moves or other details) for use within and between most computer programs in this domain.

## III. Arrays vs. Strings

In most computer chess software, the chessboard is usually represented as a (conceptually) two-dimensional (2D) array with integers to represent the 64 squares as shown in Fig. 2. Internally, however, all arrays are ultimately one-dimensional and are treated as such.

| | | | | | | | |
|---|---|---|---|---|---|---|---|
| 0 | 1 | 2 | 3 | 4 | 5 | 6 | 7 |
| 8 | 9 | 10 | 11 | 12 | 13 | 14 | 15 |
| 16 | 17 | 18 | 19 | 20 | 21 | 22 | 23 |
| 24 | 25 | 26 | 27 | 28 | 29 | 30 | 31 |
| 32 | 33 | 34 | 35 | 36 | 37 | 38 | 39 |
| 40 | 41 | 42 | 43 | 44 | 45 | 46 | 47 |
| 48 | 49 | 50 | 51 | 52 | 53 | 54 | 55 |
| 56 | 57 | 58 | 59 | 60 | 61 | 62 | 63 |

Fig. 2. Typical array representation of the chessboard.

One can imagine how all the squares of a chessboard correspond to such an array with the piece type stored in memory and referenced using an array index. For instance, in an array allocated with 64 memory locations (indices 0-63), *chessboard[9]= 'b'* can be a command to indicate that the '*b7*' square on the board contains a black bishop. A FEN string can therefore easily be derived from an array of any position along with some additional information such as the side to move, castling permissions and a potential en passant capture square. An issue with this approach may be that passing around arrays like this many thousands or even millions of times in a program uses significant resources.

Arrays are also usually passed between functions and subroutines mainly 'by reference' which is risky in the sense that the original array (i.e. its memory locations), not unlike a global variable, could be accidentally altered or lost at any time. On the other hand, many modern programming languages tend to be quite highly optimized for string manipulation as well. These tend to be passed around as a 'copy', at the expense of some speed, but with less risk to the original memory locations. It is difficult to say if arrays or strings are more efficient given a particular task as they are dependent on many other factors such as the processor, memory, compiler settings and the code itself [12].

In general, and also based on some of our own rudimentary testing with regard to chess positions, arrays are probably around 15% faster than strings. Noteworthy is that some programming languages may, on some level, even treat strings as 'character arrays'. In any case, the relevance of the algorithm presented in this article is with regard to the typical tasks of, 1) receiving a position as a FEN string (e.g. from an external program) and a move to process, 2) processing that move in the array, and then 3) outputting an updated FEN string. An array here serves as an (additional) intermediary step between the original FEN string received and the updated FEN string (after the move).

The algorithm presented in this article allows a program to achieve the same thing without the use of any intermediary array. The processing of the move is performed 'directly' on the received FEN to create the updated FEN for output. There is no conversion of FEN to array and back to FEN. Logically, this is more efficient. We do not suggest FEN strings be used for most internal purposes in computer programs because there are likely times when arrays could prove to be more efficient (even though perhaps riskier) but it may be prudent in some cases to use a combination of both the traditional array processing approach and the proposed algorithm for both internal and external processes in a program. We could not find any relevant computational research in this area.

## IV. The Algorithm Explained

We present the algorithm as a function that accepts two parameters (i.e. the FEN string, the move to be made) and returns the updated FEN string, i.e. after the move is played. The move may be represented using the algebraic notation, e.g. '*e2e4*', '*b8c6*'. Sometimes a hyphen or dash is used between the first and last two characters. The first two represent the location of the piece about to move and the last two, its destination square. The basic idea is that only a few characters in the original FEN string need to be modified to reflect the position after the move. Consider the following FEN string, the move to be made and the updated FEN, as shown in Table I.

TABLE I. FEN STRING BEFORE AND AFTER A MOVE (CASE I)

| | |
|---|---|
| FEN Before | 7N/**1b3RN1/7k**/6b1/KBp4p/5q2/6Q1/7n w - - 0 1 |
| *the move to be made: 'f7f6'* | |
| FEN After | 7N/**1b4N1/5R1k**/6b1/KBp4p/5q2/6Q1/7n b - - 0 1 |

The white rook on '*f7*' has moved to the '*f6*' square (i.e. along the 'f' file) and this change affects ranks 7 and 6 (in bold). An inversion of the 'w' to 'b' indicates it is now Black's move. It is easy to determine which segments, separated by a slash, the move played affects given that the first segment of the FEN, from the left, refers to rank 8, the second refers to rank 7 etc. So a move like '*f7f6*' can only affect segments 2 and 3. Consider another example but a different move as shown in Table II.

TABLE II. FEN STRING BEFORE AND AFTER A MOVE (CASE II)

| | |
|---|---|
| FEN Before | 7N/**1b3RN1**/7k/6b1/KBp4p/5q2/6Q1/7n w - - 0 1 |
| *the move to be made: 'f7c7'* | |
| FEN After | 7N/**1bR3N1**/7k/6b1/KBp4p/5q2/6Q1/7n b - - 0 1 |

The white rook on '*f7*' in this case has moved to the '*c7*' square (along the $7^{\text{th}}$ rank) but this change only affects *one* rank (in bold), and an inversion of the 'w' to 'b'. Rather than generating an array representation of the board, making changes to it to reflect the move, and then generating an updated FEN based on the updated array to reflect the new position (i.e. reading the whole array from upper left to lower right), only the relevant segment(s) of the string need to be manipulated based on the move in order to produce the updated FEN.

In this way, even an entire game can be played out from the starting position FEN without needing to fill any board arrays until (and only if) it is necessary to display the position on a graphical board. Let us now consider how segments can be modified directly based on the move. Consider 'Case II' above and assume that each integer in the original segment is 'expanded' to something equivalent using single digits of '1'. The segment in question would appear as in Table III with the more precise changes in bold, i.e. '111R' changes to 'R111'.

TABLE III. ONE SEGMENT MODIFICATION (SAME RANK)

| | | | | | | | | | |
|---|---|---|---|---|---|---|---|---|---|
| Before Move | Original | 1 | B | 3 | R | N | 1 | | |
| | Equivalent | 1 | B | **1** | **1** | **1** | **R** | N | 1 |
| *Rook on f7 to c7 square (same rank, different file)* | | | | | | | | | |
| After Move | Original | 1 | B | R | 3 | N | 1 | | |
| | Equivalent | 1 | B | **R** | **1** | **1** | **1** | N | 1 |

The expanded segment after the move has the 'R' shifted three places to the left because the white rook has moved three steps to the left. This is due to the difference in 'files' between the sixth file 'f' and the third file 'c', i.e. *a, b, c, d, e, f, g, h*. The 'R' (or whichever piece happens to be there) needs to vacate its location, leaving an empty square or '1' and occupy whatever square is three places to the left. If the landing square is blank, its '1' is replaced by the piece code or 'R', in this case. Even if there was another piece there, it would be replaced by 'R' because that piece is removed from the board.

Similarly, when two segments need to be changed (e.g. when moving from one rank to another or along the same file), the piece that vacates one segment is replaced by a '1' in its slot, whereas the segment that receives the piece has its corresponding slot (based on the target square's file) filled by that piece. As shown in Table IV, the 'R' leaves the '1b111RN1' (location) segment turning it into '1b1111N1' and 'lands into' the corresponding slot (same 'f' file in this case) in the (destination) segment shown below it, changing the '1111111k' into '11111R1k'. The more precise changes are in bold. The segment the rook left now can be compacted into '1b4N1' and the segment it landed into compacted into '5R1k'. The consecutive '1s' are recombined into a whole integer to be compatible with the FEN format.

TABLE IV. TWO SEGMENT MODIFICATION (DIFFERENT RANK)

| | | | | | | | | | |
|---|---|---|---|---|---|---|---|---|---|
| Before Move | Original | 1 | b | 3 | R | N | 1 | | |
| | Equivalent | 1 | b | 1 | 1 | 1 | **R** | **N** | **1** |
| *Rook on f7 to f6 square (different rank, same file)* | | | | | | | | | |
| After Move | Original | 1 | b | 4 | N | 1 | | | |
| | Equivalent | 1 | b | 1 | 1 | 1 | **1** | **N** | **1** |
| Before Move | Original | 7 | k | | | | | | |
| | Equivalent | 1 | 1 | 1 | 1 | 1 | **1** | **1** | **K** |
| *Rook on f7 to f6 square (different rank, same file)* | | | | | | | | | |
| After Move | Original | 5 | R | 1 | k | | | | |
| | Equivalent | 1 | 1 | 1 | 1 | 1 | **R** | **1** | **K** |

This expansion of (at most) two FEN segments, the replacement of particular characters and the contraction back into whole integers allows the quick and efficient string manipulation of the original FEN to reflect what the position would look like after the move has been made. A 'diagonal' move which results in a different rank *and* different file (e.g. '*b8c6*') will also require the modification of just two segments. When this is done, the remaining portion of the FEN, i.e. the 'w - - 0 1' component will need to be modified slightly as well.

As mentioned earlier, whose move it now is, i.e. 'w' or 'b', will need to be simply inverted. Existing castling rights can be copied from the original FEN if the move did not involve any of the related kings or rooks. It is easy to determine if a 'K' or 'r' happens to be among the characters in a segment that requires change and the related castling right can simply be omitted in the updated FEN, if need be. An example might be 'KQkq' becoming 'Qkq' meaning the white king can no longer castle on the king's side after the move. The king's side of the board is toward the right from White's perspective. A move may prevent castling on the next move (e.g. the king cannot pass over a square attacked by an enemy piece) but this does not negate its *right* to still do so at some future point.

The en passant square, if any, will no longer be valid if such a capture was not made immediately in the provided move. The updated FEN should therefore reflect this change. A two-square pawn move may, however, create a new en passant square. This can be easily derived from the move provided if it lands into a segment and creates either the sequences, 'Pp' or 'pP' (neighboring enemy pawns). If so, the square 'behind' the pawn that just moved, i.e. the target capture square, should be marked in the updated FEN as the en passant square. If the provided move is say, '*e7e5*' and the possibility of en passant now exists because it created a 'pP' or 'Pp' sequence in rank 5 (fourth segment from the left), the target square would simply be '*e6*', i.e. behind the '*e5*' square.

The 'halfmove' may be copied directly from the original FEN without modification if the position is to be treated independently (even after the move) since it is not relevant to any situation outside of an actual game with a sequence of moves occurring in a specific order. The 'fullmove' should be incremented by one after White makes a move. If the position is *not* treated independently (e.g. part of a game), then a record of the moves played should also be kept. With chess compositions (i.e. problems, puzzles, constructs), for instance, these two fields are not particularly relevant since only the initial position is critical.

If more is required such as the actual moves played (in a game), the solution (to a composition), variations, commentary etc., these can be stored along with the FEN string in a PGN (portable game notation) file. The PGN format was also developed by Steven J. Edwards around 1993 [13]. Unlike the FEN and EPD formats which are each used to describe a single chess position (the latter being more detailed, as explained in Section II), a PGN can store one or more positions and any associated moves (including variations, commentary etc.). In short, a database of chess games or compositions. It is widely used even today by modern chess programs such as ChessBase [14].

The last issue is that of pawn promotions, i.e. to queen, rook, bishop or knight. Typically the notation might look like '*c7c8Q*' or '*b2b1N*'. Promotion moves can easily be identified because the piece moving is a pawn and the target rank is only either rank 8 (for White) or rank 1 (for Black). This would mean that the pawn (i.e. 'P' or 'p') in the location segment is replaced by '1' and the corresponding slot in the destination segment is replaced by whatever the promotion piece specified in the move happens to be, i.e. 'Q', 'R', 'B' or 'N'. The steps just explained can be summarized as follows.

1. Accept the original FEN (OFEN) and move to be made (MTBM).
2. Separate MTBM into location (LOC) and destination (DES); also promotion piece (PP), if applicable.
3. Identify segment(s) in OFEN pertaining to LOC and DES.
4. Expand integers in segment(s) to component '1s'.
5. Identify the relevant characters in LOC and DES.
6. Remove piece character from LOC segment and replace with '1'.
7. Replace '1' or piece character in DES with LOC piece character or PP character, if applicable.
8. Contract segment(s) so '1s' become whole integers again.
9. Invert the side to move.
10. Append or modify the castling permissions, if applicable.
11. Append or modify the en passant square, if applicable.
12. Append or modify the 'halfmove' and 'fullmove' values, if applicable.

Depending on the programming language used, the actual implementation of the algorithm in code may involve more steps within each described above but the process of character string handling should be equivalent. The updated FEN should therefore be immediately ready for further internal use or export to any other program that can read the notation. Many command line or console-based chess engines (e.g. Stockfish) receive a position directly in the form of a FEN string in order to solve for mate or find the best move to play. The graphical or visual board is treated as an optional third-party extension.

## V. Some Examples

Consider the constructed positions shown in Figs. 3-5. The new FEN in each has the changed segments in bold. Fig. 3 is a typical position with one castling permission. Fig. 4 results in an en passant square after the move. Fig. 5 shows a promotion to knight after the move. Where there are multiple '1s' next to each other, they can be identified based on their position from left to right, e.g. 'second', 'fifth'. The same or similar identification method can be applied to multiple pieces or pawns next to each other, e.g. 'ppp', 'bbnn'. A blank rank with others next to it in the FEN (e.g. '8/8/8') may be individually identified using the actual rank number such as 'rank 8' (i.e. on the far left) or 'rank 6'.

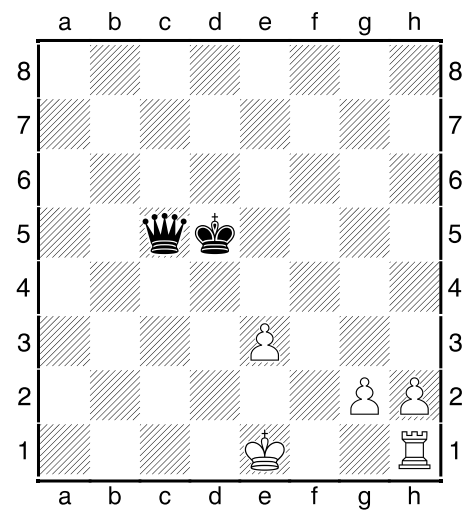

1. 8/8/8/2qk4/8/4P3/6PP/4K2R w K - 0 1, 'e1e2'
2. LOC: 'e1', DES: 'e2'
3. LOC: '4K2R', DES: '6PP'
4. LOC: '1111K11R', DES: '111111PP'
5. LOC: 'K', DES: '1' (fifth)
6. LOC: '1111111R'
7. DES: '1111K1PP'
8. LOC: '7R', DES: '4K1PP'
9. 'w' = 'b'
10. 'K' = '-'
11. Not applicable
12. '0 1' = '0 2'

Updated FEN:
8/8/8/2qk4/8/4P3/**4K1PP/7R** b - - 0 2

Fig. 3. Example position A run through the algorithm.

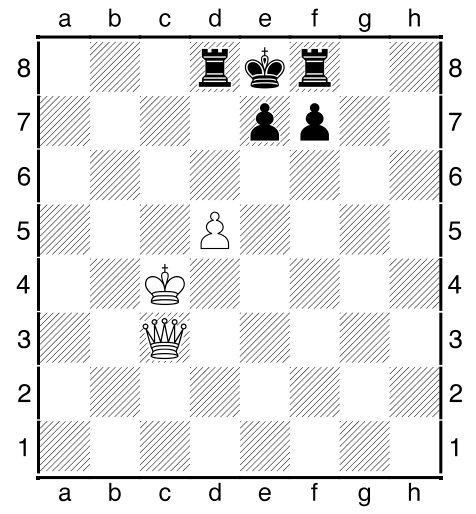

1. 3rkr2/4pp2/8/3P4/2K5/2Q5/8/8 b - - 0 1, 'e7e5'
2. LOC: 'e7', DES: 'e5'
3. LOC: '4pp2', DES: '3P4'
4. LOC: '1111pp11', DES: '111P1111'
5. LOC: 'p' (first), DES: '1' (fourth)
6. LOC: '11111p11'
7. DES: '111Pp111'
8. LOC: '5p2', DES: '3Pp3'
9. 'b' = 'w'
10. Not applicable
11. 'e6' (i.e. behind 'e5')
12. '0 1' = '0 2'

Updated FEN:
3rkr2/**5p2**/8/**3Pp3**/2K5/2Q5/8/8 w - e6 0 2

Fig. 4. Example position B run through the algorithm.

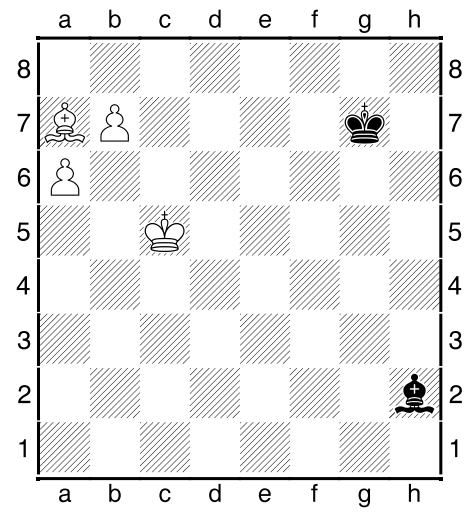

1. 8/BP4k1/P7/2K5/8/8/7b/8 w - - 0 1, 'b7b8N'
2. LOC: 'b7', DES: 'b8', PP: 'N'
3. LOC: 'BP4k1', DES: '8' (rank 8)
4. LOC: 'BP1111k1', DES: '11111111'
5. LOC: 'P', DES: '1' (second)
6. LOC: 'B11111k1'
7. DES: '1N111111'
8. LOC: 'B5k1', DES: '1N6'
9. 'w' = 'b'
10. Not applicable
11. Not applicable
12. '0 1' = '0 2'

Updated FEN:
**1N6/B5k1**/P7/2K5/8/8/7b/8 b - - 0 2

Fig. 5. Example position C run through the algorithm.

## VI. Conclusions

The FEN notation has been around in some form or other for nearly 140 years and proved to be very useful for recording individual chess positions. It is unfortunate that for most of recorded chess history (which spans over a thousand years [15]), there was likely no established means of recording positions except perhaps to draw them out graphically somehow. With approximately $10^{46}$ possible unique positions [16], the vast majority of chess positions will never be known by humans. This is why those that stand out or are appealing (e.g. some compositions, certain points in certain games) should be able to be easily recorded and shared for posterity.

An interesting game or composition is often talked and written about for decades, if not centuries [17-19]. With the advent of very strong (both free and commercially available) chess engines, even the moves to play given any position have become less relevant since the engines can, at least in most cases, easily derive them. Of course between human players (even grandmasters), the moves actually played from a given position may not always be the best. In fact, the moves are often affected by their personalities and playing styles [20] which is what also makes them interesting to people and worth recording.

Therefore a FEN string along with the actual moves played, variations considered and even commentary by certain people all within a PGN file or database may be preferable overall. In this article, we focused only on the FEN which is the most critical component when recording a chess position. Until now, the typical practice has been to rely on an array of some kind to store the piece locations (read from a FEN string), process any moves attached, and then generate an updated FEN by scanning said array. We proposed an algorithm that bypasses any need for an intermediary array and can process individual moves directly into an updated FEN, which is immediately ready for export.

While this may not be more efficient than arrays for internal processing in chess software, it can be used alongside arrays for side or additional processing of a position or node without affecting the primary array memory locations (or calling for new arrays to be generated). For instance, in evaluating side variations (for commentary [21]) or in determining the aesthetic value of a position [22] during automatic composition, i.e. something secondarily related to the primary task being performed on the position.

The clear examples provided along with the algorithm demonstrate that there is no position or move (including involving castling, en passant or pawn promotion) that cannot be correctly processed using this approach. Further work may include testing this algorithm with FEN or FEN-like notation for chess variants and other board games. Having a solely string-based alternative (to arrays) in processing board game positions and moves is in itself undeniably beneficial in these domains.

## APPENDIX A

The following italicized text in quotes, sourced from [2] as cited in [6], was adjusted slightly for paragraphing and a modern graphic board added to better illustrate the position. It was written within David Forsyth's lifetime and describes the notation format with regard to the man himself.

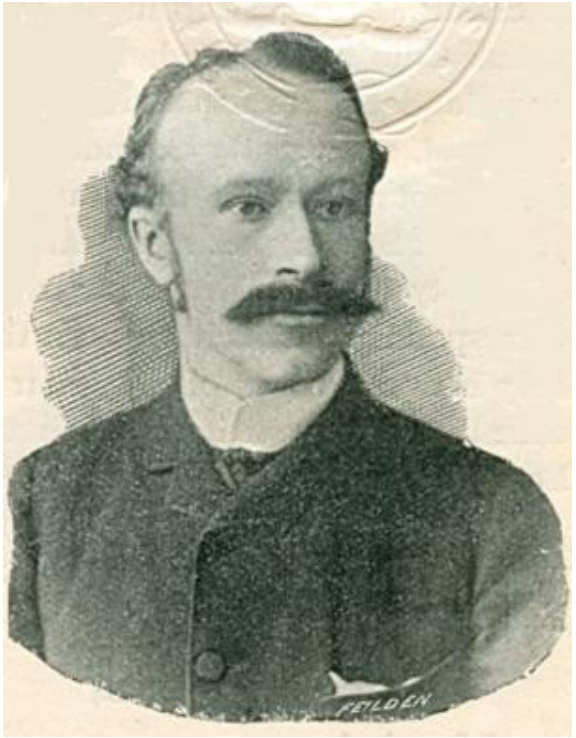

Fig. 6. David Forsyth [23].

"*It was only in his twenty-sixth year that he learnt chess. Prior to then he was totally ignorant of anything connected with the game, but, like most persons who have an aptitude for chess, he rapidly attained great strength, as is shown by his appearance year or two afterwards (in July, 1884) in the Major Tournament of the first Congress of the Scottish Chess Association at Glasgow, where he won majority of games against the strongest and most experienced players in Scotland, and for time seemed first favourite for the championship. Mr. Forsyth possesses in marked degree the rare faculty of blindfold play.*

*In reference to the invention of the Forsyth notation, we are informed that, at the earliest stage of his chess career, he had intuitively used for recording chess positions notation of his own, which, so far as he suspected, possessed neither originality nor utility. On showing it to some chess friends they pronounced it eminently useful and in order that the chess public might use it if they chose, he gave an explanation of it in the Glasgow Weekly Herald of February 10th, 1883. It has since been explained in many works on chess, among others in Steinitz's Modern Chess Instructor, Mason's Principles of Chess, and Rowland's Problem Art.*

*The notation is very simple, concise, and extremely useful for taking down end-games and positions of adjourned games. White pieces are denoted by capitals, the black by small letters, or the black pieces can be underlined to distinguish them from the white. Place the board before you as if playing the white pieces, and begin counting from the top left-hand corner (Black's QR sq.), and put down the number of squares which are empty till piece or pawn is reached, always counting any rank from the left side of the board. When piece or pawn is reached place its name as written by capital or small letters according as it is white or black piece, and continue till White's KR sq. is reached. All empty squares are thus denoted by numerals, while occupied squares will be indicated by letters. Problem No. would be represented thus*

*1 B 6, 2 kt 5, p 1 Kt 1 P 2 R, P 1 K 3 Kt 1, 4 P k 2, 1 Q 2 p 2 p, 6 kt P, 1 B 4 R 1.*

*(White mates in two)*

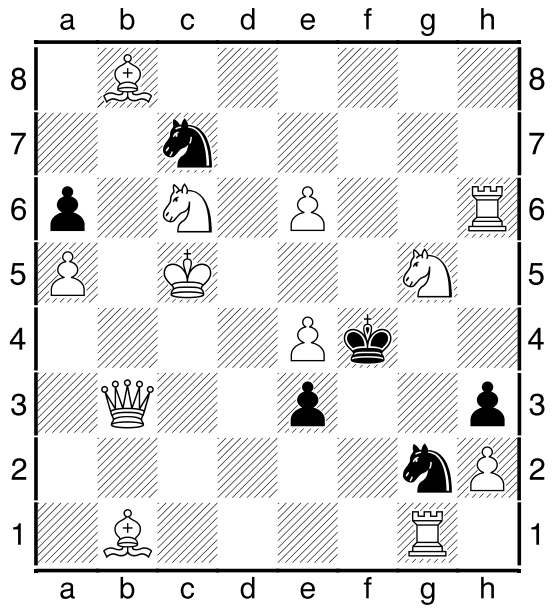

*(Mrs. W. J. Baird composition)*

*Mr. Forsyth modestly overlooks the great service he rendered to chess in inventing this notation, and rests his principal claim to usefulness in the sphere of chess on his qualities as an organiser. Shortly after joining the Glasgow Chess Club in 1883 he was appointed Secretary, and afterwards treasurer, both of which offices he resigned on removing to Edinburgh in 1887. He has been secretary and treasurer of the Scottish Chess Association since its inauguration in 1884.*

*For several years prior to his removal from Glasgow, he assisted in editing the Glasgow Weekly Herald chess column, and since November 4th, 1893, he has conducted the well known column in the Weekly Scotsman. He avoids all gossipy or controversial matter, or attempts at wit, or poetry. Favouring no nationality or clique of players, but judging all by genuine merit, Jew, Gentile or Mahommedan, Mr. Forsyth can justly claim that his column is conducted on cosmopolitan lines."*

Even though there may have been some objections around that time regarding the usefulness of the notation, in that it did not record the moves or indicate any particular square [6], the notation fairly quickly evolved somewhat, i.e. with the use of slashes ('/') between ranks. We could not determine whom, exactly, proposed the use of the slash but reputedly both Mr. Forsyth and one Mr. Rayner, in an exchange of letters, agreed it was a good idea (ibid.).